\RequirePackage{fix-cm}
\RequirePackage{rotating}

\documentclass[twocolumn]{svjour3}          
\smartqed  
\usepackage{graphicx,url,xcolor}

\usepackage{todonotes} 
\usepackage{xurl}
\usepackage[sort,numbers]{natbib}
\usepackage{rotating,hyperref,balance}
\usepackage{multirow}

\usepackage{tikz}
\usepackage{subcaption}
\usetikzlibrary{automata,positioning}

\setcitestyle{square}

\usepackage{pgfplots}

\graphicspath{{./figures/}}
\newcommand*{\affaddr}[1]{#1} 
\newcommand*{\affmark}[1][*]{\textsuperscript{#1}}

\newcommand{\jialin}[1]{{\textcolor{blue}{#1}}}
\newcommand{\revised}[1]{{\textcolor{black}{#1}}}
\begin{document}
\sloppy

\title{Deep Learning for Procedural Content Generation
}


\author{Jialin Liu\affmark[1] \and
        Sam Snodgrass\affmark[2] \and
        Ahmed Khalifa\affmark[3]\and 
        Sebastian Risi\affmark[2,4]\and
        Georgios N. Yannakakis\affmark[2,5,6]\and
        Julian Togelius\affmark[2,3]
}

\authorrunning{J. Liu, S. Snodgrass, A. Khalifa, S. Risi, G. N. Yannakakis and J. Togelius} 

\institute{J. Liu\at
              \email{liujl@sustech.edu.cn} 
           \and
           S. Snodgrass\at
              \email{sam@modl.ai}
              \and
            A. Khalifa\at
              \email{ahmed@akhalifa.com} 
              \and
            S. Risi \at
              \email{sebr@itu.dk}  \and
            G. N. Yannakakis\at
            \email{georgios.yannakakis@um.edu.mt} 
            \and
            J. Togelius\at
             \email{julian@togelius.com} \\
              ~\\
        \affaddr{\affmark[1]Guangdong Provincial Key Laboratory of Brain-inspired Intelligent Computation, Department of Computer Science and Engineering, Southern University of Science and Technology, Shenzhen, China}\\
        \affaddr{\affmark[2]Modl.ai, Copenhagen, Denmark}\\
       \affaddr{\affmark[3]New York University, New York, USA}\\
       \affaddr{\affmark[4]IT University of Copenhagen, Copenhagen, Denmark}\\
        \affaddr{\affmark[5]Institute of Digital Games, University of Malta, Msida, Malta}\\
        \affaddr{\affmark[6]Technical University of Crete, Chania, Greece}
}

\date{Received: date / Accepted: date}

\maketitle

\begin{abstract}

Procedural content generation in video games has a long history. Existing procedural content generation methods, such as search-based, solver-based, rule-based and grammar-based methods have been applied to various content types such as levels, maps, character models, and textures. A research field centered on content generation in games has existed for more than a decade. More recently, deep learning has powered a remarkable range of inventions in content production, which are applicable to games. While some cutting-edge deep learning methods are applied on their own, others are applied in combination with more traditional methods, or in an interactive setting. This article surveys the various deep learning methods that have been applied to generate game content directly or indirectly, discusses deep learning methods that could be used for content generation purposes but are rarely used today, and envisages some limitations and potential future directions of deep learning for procedural content generation.
\end{abstract}



\section{Introduction}\label{intro}

Deep learning has powered a remarkable range of inventions in content production in recent years, including new methods for generating audio, images, 3D objects, network layouts, and other content types across a range of domains. It stands to reason that many of these inventions would be applicable to games. In particular, modern video games require large quantities of high-definition media, which could potentially be generated through deep learning approaches. For example, promising recent methods for generating photo-realistic faces could be used for character creation in games.

At the same time, video games have a long tradition of procedural content generation (PCG)~\citep{togelius2011procedural}, where some forms of game content have been generated algorithmically for a long time; the history of digital PCG in games stretches back four decades. \revised{In the last decade and a half, we have additionally seen a research community spring up around challenges posed by game content generation~\citep{togelius2011search,togelius2013procedural,shaker2016procedural,yannakakis2018artificial,risi2019increasing,summerville2018procedural,Kegel2020Puzzle}.} This research community has applied methods from core computer science, such as grammar expansion~\citep{dormans2010adventures}; AI, such as constraint solving~\citep{smith2011answer} and evolutionary computation~\citep{browne2010evolutionary,togelius2011search}; and graphics, such as fractal noise~\citep{ebert2003texturing}. But only in the last few years have we seen a real effort to bring the tools of deep learning to game content generation.





Deep learning brings new opportunities and leads to exciting advances in PCG, such as generative adversarial networks  (GANs)~\citep{goodfellow2014GAN}, deep variational autoencoders (VAEs)~\citep{Kingma2013vae} and long short-term memory (LSTM)~\citep{hochreiter1997lstm,Greff2017lstm}.
However, those methods for other generative or creative purposes are not always applicable to games and need certain adaptations due to the functionality criteria of different game content. Methods for generating images (e.g., generative networks) can be used to generate image-like game content (e.g., level maps, landscapes, and sprites). However, the generated levels should be playable and require specific gameplay skill-depth. The generated sprites should imply specific character or emotion, as well as coherence within the game. Training reliable models requires a necessary amount and quality of data, while the available data of content and playing experience for most games is limited. Careful consideration and sophisticated design of adaptation techniques are requisites for applying deep learning methods to generate game content.

It is important to note that content generation has uses outside of designing and developing games for humans to experience. In addition to creating content in games meant for humans to play, content generation can also play a crucial role \revised{in creating generalizable game-based and game-like benchmarks for reinforcement learning and other forms of AI~\citep{torrado2018deep,fang2020adaptive}.}


This article surveys the various approaches that have been taken to generate game content with deep learning, and also discusses methods proposed from within deep learning research that could be used for PCG purposes. First, we give an overview of types of game content that could conceivably be generated by deep learning, including the particular constraints and affordances of each content type and examples of such applications (if they exist), followed by an overview of applicable deep learning methods. 

\section{Scope of The Review}\label{sec:scope}

\revised{This article discusses the use of deep learning (DL) methods, here defined as neural networks with at least two layers and some nonlinearity~\cite{goodfellow2016deep}, for game content generation.} We take an inclusive view of games as any games a human would conceivably play, including board games, card games, and any type of video games, such as arcade games, role-playing games, first-person shooters, puzzle games, and many others. 
\revised{Several other surveys and overviews of PCG in games already exist. Here, we delineate the scope of our article by comparing it to existing books and surveys in Section \ref{sec:sota} and Section \ref{sec:novelty}. Section \ref{sec:selection} describes our paper selection methodology.}

\subsection{Related Work}\label{sec:sota}

\revised{A number of books and surveys of PCG with different focuses and aims have been published in the past two decade~\citep{togelius2011search,togelius2013procedural,shaker2016procedural,yannakakis2018artificial,risi2019increasing,summerville2018procedural,Kegel2020Puzzle}.}
The two textbooks for PCG~\cite{shaker2016procedural} and Game AI~\cite{yannakakis2018artificial} cover the search-based methods, solver-based methods, constructive generation methods (such as cellular automata and grammar-based methods), fractals, noise, and ad-hoc methods for generating diverse game content. \revised{\citet{Kegel2020Puzzle} reviewed the PCG methods for eleven categories of puzzles, but few work based on deep learning has been reported.}
The article by \citeauthor{togelius2011search} reviews the search-based PCG methods, defined as using meta-heuristics to search in a predefined content space, not necessarily represented by the same format of the content itself, and automatically generate new content~\citep{togelius2011search}. The search is led by a fitness or evaluation function which measures the quality or playability of the generated content. The experience-driven PCG framework \citep{yannakakis2011experience} largely adopts a search-based approach and reviews ways in which algorithms can generate content for adjusting the player experience. Most of the reviewed search-based methods in both survey papers rely on evolutionary algorithms. In this article, we also cover some search-based methods which cooperated with deep learning methods for generating content. The most famous example may be latent variable evolution~\citep{bontrager2018LVE}.
\citet{risi2019increasing} focuses on PCG for applications in Reinforcement Learning (RL), while the work based on RL methods reviewed in this article mainly used RL agents to play the generated levels, which indirectly served as content evaluators. \citet{khalifa2020pcgrl} models the level generation as an iterative process that one needs to edit the levels to meet certain requirements or achieve some specific goals. RL agents need to learn to generate levels through this iterative process.
\revised{The study of \citet{summerville2018procedural}, published in 2018, reviews the PCG via Machine Learning (PCGML) methods, building on e.g. Markov chains (e.g.,~\cite{Summerville2015MCMCTSP4,snodgrass2015hierarchical,snodgrass2016controllable,snodgrass2016learning,zafar2019generating}), n-grams~(e.g., \cite{dahlskog2014linear}), and Bayes nets~(e.g.,\cite{guzdial2016game}), whereas we will focus exclusively on deep learning in this article.}

\subsection{Novelty of The Review}\label{sec:novelty}

\revised{The differences between the current article and the PCGML survey~\cite{summerville2018procedural} is that (i) our article focuses on DL-based methods, defined at the beginning of Section \ref{sec:scope} (although other techniques will be mentioned for contrast); (ii) our article surveys more types of game content, such as narrative text and graphical textures; (iii) we also discuss applications of deep learning to support PCG, such as for content quality prediction; and (iv) our survey is written more than three years after the PCGML survey was first submitted and two years after it was published, during which time an avalanche of new work in the field has appeared.}

During the two years after the publication of \cite{summerville2018procedural}, PCG via deep learning has been growing quickly and a significant number of papers and articles have been published. The trend was mainly set by latent variable evolution~\citep{bontrager2018LVE} in 2018. A review of the state-of-the-art and the latest applications of deep learning to PCG is needed. 

\subsection{Paper Collection Methodology}\label{sec:selection}

\revised{To collect the related papers published or online since 2018, till end of August 2020, we have searched with Google Scholar and Web of Science using the following search terms \emph{( ``game'') AND (``design'')} and \emph{
(``game'') AND (``procedural content generation'' OR ``pcg'')}, separately.
We systematically went through the returned papers, most of which were publications in the IEEE Transactions on Computational Intelligence and AI in Games (T-CIAIG), the IEEE Transactions on Games (ToG), in the proceedings of the IEEE Conference on Computational Intelligence and Games (CIG) series, the IEEE Conference on Games (CoG) series, the International Conference on the Foundations of Digital Games (FDG) series, the Artificial Intelligence for Interactive Digital Entertainment (AIIDE) Conference series and their related workshops, as well as special sessions at other conferences, such as the IEEE Congress on Evolutionary Computation (IEEE CEC). 
We also went through the papers that have been recently accepted in 2020 by the conferences mentioned above.
Only work that involve direct or indirect use of DL-based methods for generating game content or evaluating content or content generators are reviewed in this article, while the ones being returned due to citations with the search terms but are out of our scope are not included.
}

\section{Content Types}\label{sec:contenttypes}

Generally, game content can be distinguished from the content meant for non-interactive media by various forms of functionality constraints. Video, images, and music all require coherence, and in general that aesthetic suffers when the coherence fails. For example, GANs can often create images that are locally convincing but globally incoherent, such as a side-view of a car where the front wheels have a different size and style to the back wheels. This may be annoying to the human viewer, but the image still unmistakably depicts a car; it doesn't turn into a blur of random pixels just because the wheels on the car don't match. In contrast, when generating a game level, if the final door has no matching key the level is unplayable; the level's utility as content is not just slightly diminished, but essentially zero (unless manually repaired). Making a neural network learn to produce only functional content is often a tall task, and is one of the core challenges of using deep learning for PCG. Not all types of game content have the same extent of functional constraints however, and some offer affordances that may make content generation relatively easier. Also, not all content is \emph{necessary}; depending on the game's design, there might be artifacts that are allowed to be broken, as the user can simply discard them and select others. Weapons in \emph{Borderlands} are a good example of optional content.

\subsection{Game Levels}
The most common type of content to generate in games is levels. These are spaces in two or three dimensions that need to be traversed. Typically, these are necessary rather than optional, and have strong functional constraints that require them to be playable. For example, there can not be impassable geometry (such as gaps or walls) blocking traversal of the level, items needed to finish the levels must be present, and enemies cannot be unbeatable. 2D, side-scrolling platform games is a genre where procedural generation is particularly common, both in entertainment-focused games (in particular indie games) and in academic research. Among the former, the standout game Spelunky has defined a way of building 2D platform games around PCG; among the latter, the Mario AI Framework\revised{~\cite{togelius2013championship}}, built around an open-source clone of Super Mario Bros, has been used in so many research projects that it could be called the ``drosophila of PCG research''. Another type of commonly attempted 2D level is the rogue-like or dungeon-crawler level, where the objectives and constraints are similar to the platform game level, but which are viewed from the top down so physics works differently. Related to this are levels for first-person shooters.
Another kind of 2D level is the battle map, used in strategy games such as StarCraft or player-versus-player modes of first-person shooters. While such maps also have ``hard'' constraints, such as sufficient room for the players' bases, there are also the softer constraints of balancing; many features contribute to the quality of battle maps, but balancing is paramount. 

Levels for music games, such as as Guitar Hero or Dance Dance revolution, can be seen as 2D levels as well. Here the player is automatically moved along the level, and has to carry out certain actions in time with the music, as prompted by level features. Some interesting work has been done on learning to create such music game levels from existing music \revised{(e.g., \cite{tsujino2018dance,donahue2017dance}).}

\subsection{Text}
Almost all games include some form of text, and typically they use text to convey narrative. This text typically has very strong constraints, as it needs to be truthful with regards to what happens in the game. For example, if the text says that the King lives in Stockholm, this must actually be the case lest it misleads the player. Traditionally, generative text in games has not been very ambitious and used simple text substitution or grammar-based approaches. Outside of games, deep learning has made great strides with LSTM networks~\cite{hochreiter1997lstm,Greff2017lstm} and,
more recently, transformers able to generate coherent and stylistically relevant text. However, these methods are not easy to integrate into most games because of the lack of control over deep learning-based text generators. However, games such as AI Dungeon 2 have managed to build gameplay on top of almost uncontrollable text generation.

\subsection{Character Models}
Faces and character models are examples where deep learning has advanced content creation capabilities radically in recent years, but these methods have generally not made their way into games. Datasets of thousands of real human faces, such as the Celeb-A dataset\revised{~\cite{liu2015faceattributes}}, have become standard benchmark for developing new GAN variations, leading to some impressive breakthroughs in face generation. While many games have a need for (human) faces in various roles, including for freshly generated NPCs, the character design feature of role-playing games is a standout application case for controllable PCG, where machine learning-based methods have yet to make their mark. Depending on the features of the game, these faces or models might need to be animatable, so that they can produce believable movements or facial expressions.

\subsection{Textures}
Textures are used in almost all 3D games, and is perhaps the type of content that has the fewest functionality constraints. Procedural methods such as Perlin Noise \cite{perlin1985image,ebert2003texturing} have been used for texture generation in games since the birth of commercial 3D games with \emph{DOOM}. Deep learning methods for texture generation could provide a viable alternative in this case.

\subsection{Music and Sound}
Most games feature a soundtrack, often composed of both music and sound effects. The constraints on the soundtrack tend to be relatively soft compared to other types of content constraints; the sound effects should be appropriate to the actions in the game at any given moment, and the music to the emotional tone of the moment, but inappropriate sound does not necessarily break the game. Quite a few games involve some kind of procedural soundtrack, and some research projects have focused on music generation able to adapt to affective shifts in real-time~\cite{scirea2018evolving}. At the same time, deep learning has made impressive strides in learning to generate music with some modes of controllability~\cite{dhariwal2020jukebox}, but we have yet to see the use of deep learning methods for sound generation in games. 

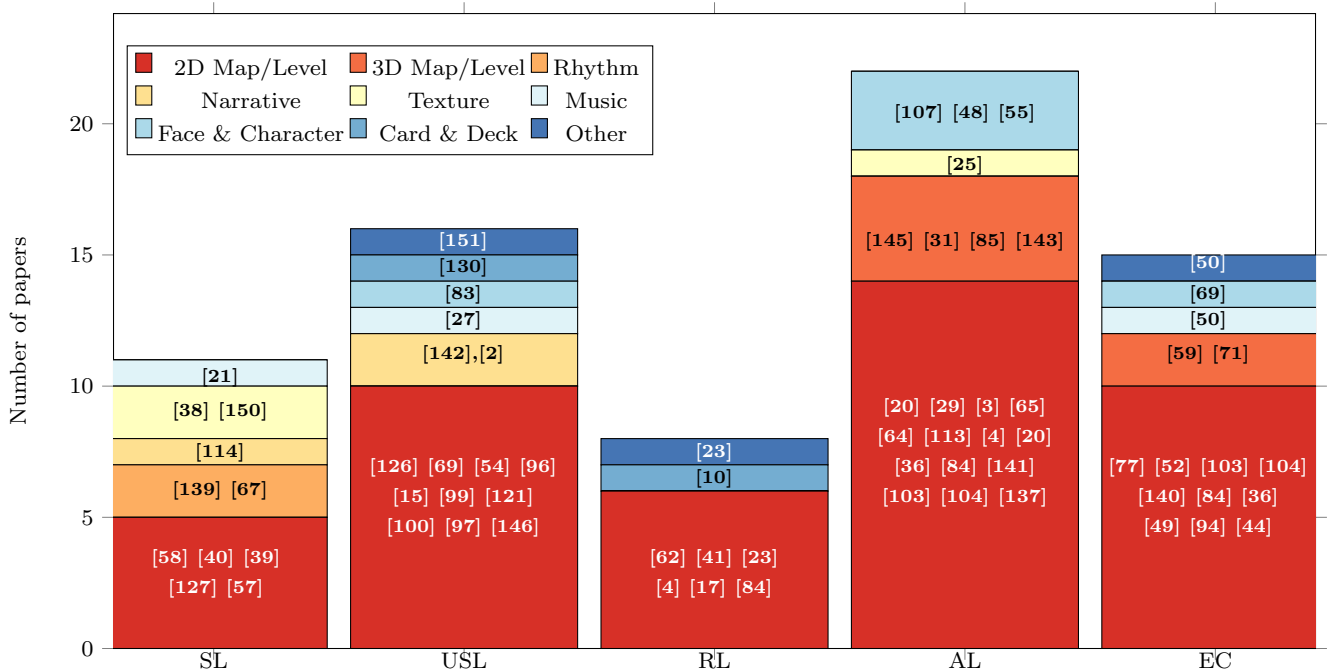
\begin{figure*}[htbp]
    \centering
\definecolor{clr1}{RGB}{215,48,39}
\definecolor{clr2}{RGB}{244,109,67}
\definecolor{clr3}{RGB}{253,174,97}
\definecolor{clr4}{RGB}{254,224,144}
\definecolor{clr5}{RGB}{255,255,191}
\definecolor{clr6}{RGB}{224,243,248}
\definecolor{clr7}{RGB}{171,217,233}
\definecolor{clr8}{RGB}{116,173,209}
\definecolor{clr9}{RGB}{69,117,180}

\begin{tikzpicture}[auto,node distance=2cm]
\begin{axis}[
    ybar stacked,
	bar width=85pt,
	width=\textwidth,
	height=10cm,
    legend style={at={(0.23,0.95)},
      anchor=north,legend columns=3},
    ylabel={Number of papers},
    ymin=0,
    symbolic x coords={SL, USL, RL, AL, EC},
    xtick=data,
    x tick label style={rotate=0,anchor=center, tick align=outside},
    ]
  \addplot+[ybar,fill=clr1,fill opacity=1,draw=black] plot coordinates {(SL,5) (USL,10) (RL,6) (AL,14) (EC,10)}; 
  \addplot+[ybar,fill=clr2,fill opacity=1,draw=black] plot coordinates {(SL,0) (USL,0) (RL,0) (AL,4) (EC,2)}; 
  \addplot+[ybar,fill=clr3,fill opacity=1,draw=black] plot coordinates {(SL,2) (USL,0) (RL,0) (AL,0) (EC,0)}; 
  \addplot+[ybar,fill=clr4,fill opacity=1,draw=black] plot coordinates {(SL,1) (USL,2) (RL,0) (AL,0) (EC,0)}; 
  \addplot+[ybar,fill=clr5,draw=black,fill opacity=1] plot coordinates {(SL,2) (USL,0)  (RL,0) (AL,1) (EC,0)}; 
  \addplot+[ybar,fill=clr6,draw=black,fill opacity=1] plot coordinates {(SL,1) (USL,1) (RL,0) (AL,0) (EC,1)}; 
  \addplot+[ybar,fill=clr7,draw=black,fill opacity=1] plot coordinates {(SL,0) (USL,1) (RL,0) (AL,3) (EC,1)}; 
  \addplot+[ybar,fill=clr8,draw=black,fill opacity=1] plot coordinates {(SL,0) (USL,1) (RL,1) (AL,0) (EC,0)}; 
  \addplot+[ybar,fill=clr9,draw=black,fill opacity=1] plot coordinates {(SL,0) (USL,1) (RL,1) (AL,0) (EC,1)}; 
\legend{\strut 2D Map/Level, \strut 3D Map/Level, \strut Rhythm, \strut Narrative, \strut Texture, \strut Music,
\strut Face \& Character, \strut Card \& Deck, \strut Other}
\end{axis}
\node at (1.4,0.8) [align=center,color=white] {\scriptsize\textbf{\cite{summerville2016learning} \cite{karavolos2017learning} 
}};
\node at (1.4,1.2) [align=center,color=white] {\scriptsize\textbf{\cite{karavolos2018pairing} \cite{guzdial2018explainable} \cite{guzdial2018co}
}};
\node at (4.6,2.4) [align=center,color=white] {\scriptsize\textbf{
\cite{summerville2016super} \cite{liapis2013transforming} \cite{jain2016autoencoders} \cite{sarkar18blending} 
}};
\node at (4.6,2.0) [align=center,color=white] {\scriptsize\textbf{
 \cite{davoodi2020approach} \cite{sarkar19controllable} \cite{snodgrass2020multi}
}};
\node at (4.6,1.6) [align=center,color=white] {\scriptsize\textbf{
 \cite{sarkar2020exploring} \cite{sarkar2020sequential} \cite{yang2020game}
}};
\node at (7.9,1.2) [align=center,color=white] {\scriptsize\textbf{
\cite{khalifa2020pcgrl} \cite{guzdial2019friend} \cite{earle2019using} 
}};
\node at (7.9,0.8) [align=center,color=white] {\scriptsize\textbf{
\cite{bontrager2020fully} \cite{delarosa2020mixed} \cite{mott2019controllable}
}};
\node at (11.2,2) [align=center,color=white] {\scriptsize\textbf{
 \cite{schrum2020interactive} \cite{schrum2020cppn} \cite{torrado2019bootstrapping}   
}};
\node at (11.2,2.4) [align=center,color=white] {\scriptsize\textbf{
 \cite{gutierrez2020zeldagan} \cite{mott2019controllable} \cite{volz2020capturing}  
}};
\node at (11.2,2.8) [align=center,color=white] {\scriptsize\textbf{
\cite{ping2020conditional} \cite{shu2020cnet} \cite{bontrager2020fully} \cite{di2020efficient}
}};
\node at (11.2,3.2) [align=center,color=white] {\scriptsize\textbf{
 \cite{di2020efficient}
        \cite{fontaine2020illuminating}
        \cite{awiszus2020toad}
        \cite{Kumaran2020Generating}
}};
\node at (14.4,1.6) [align=center,color=white] {\scriptsize\textbf{
\cite{hoover2014functional} \cite{risi2015petalz}  \cite{hastings2009automatic} 
}};
\node at (14.4,2) [align=center,color=white] {\scriptsize\textbf{
  \cite{volz2018evolving} \cite{mott2019controllable} \cite{gutierrez2020zeldagan} 
}};
\node at (14.4,2.4) [align=center,color=white] {\scriptsize\textbf{
   \cite{lucas2019tile} \cite{irfan2019evolving}  \cite{schrum2020interactive} \cite{schrum2020cppn} 
}};
\node at (11.2,5.4) [align=center] {\scriptsize\textbf{
\cite{wulff2017deep} \cite{giacomello2018doom} \cite{park2019generating} \cite{wang2020sketch}
}};
\node at (14.4,3.9) [align=center] {\scriptsize\textbf{
\cite{karavolos2018using} \cite{liapis2013sentient}  
}};
\node at (1.4,2.1) [align=center] {\scriptsize\textbf{
\cite{tsujino2018dance} \cite{liang2019procedural} 
}};
\node at (1.4,2.6) [align=center] {\scriptsize\textbf{
\cite{sirota2019towards} 
}};
\node at (4.6,3.9) [align=center] {\scriptsize\textbf{
\cite{walton2019dungeon2},\cite{ammanabrolu2020bringing} 
}};
\node at (1.4,3.15) [align=center] {\scriptsize\textbf{
\cite{guzdial2017visual} \cite{yoo2016changing} 
}};
\node at (11.2,6.4) [align=center] {\scriptsize\textbf{
\cite{fadaeddini2018case}
}};
\node at (1.4,3.6) [align=center] {\scriptsize\textbf{
\cite{donahue2017dance}
}};
\node at (4.6,4.35) [align=center] {\scriptsize\textbf{
\cite{ferreira2020computer}
}};
\node at (14.4,4.35) [align=center] {\scriptsize\textbf{
\cite{hoover2015audioinspace}  
}};
\node at (4.6,4.7) [align=center] {\scriptsize\textbf{
\cite{mordvintsev2020growing} 
}};
\node at (11.2,7.1) [align=center] {\scriptsize\textbf{
\cite{serpa2019towards} \cite{hong2019game} \cite{Jin2017Towards} 
}};
\node at (14.4,4.7) [align=center] {\scriptsize\textbf{
\cite{liapis2013transforming} 
}};
\node at (4.6,5.05) [align=center] {\scriptsize\textbf{
\cite{summerville2016mystical} 
}};
\node at (7.9,2.25) [align=center] {\scriptsize\textbf{
\cite{chen2018q}
}};
\node at (4.6,5.4) [align=center,color=white] {\scriptsize\textbf{
\cite{Yumer2015Procedural}
}};
\node at (7.9,2.6) [align=center,color=white] {\scriptsize\textbf{
\cite{earle2019using}
}};
\node at (14.4,5.1) [align=center,color=white] {\scriptsize\textbf{
\cite{hoover2015audioinspace} 
}};

\end{tikzpicture}
    \caption{This figure shows the distribution of research by methods and content types. We notice the disproportionately large amount of work on 2D level and map generation compared to all other content types. 
    }
    \label{fig:barcollection}
\end{figure*}


\section{Training Methods and Neural Architectures of DLPCG}\label{sec:methods} 

Due to the different types and roles of content in games, diverse deep learning methods have been adapted for PCG. In this section we present different ways to apply deep learning for PCG systems, the target content, and their generality. The approaches are categorized by the type of machine learning method used for training. Additionally, works \revised{combining} evolutionary computation techniques to deep learning methods are also presented.
The works reviewed in this section are summarized in Fig. \ref{fig:barcollection}, categorized by the content types and deep learning methods.

\revised{Generating different types of content often requires different types of neural architectures. In the use cases reviewed in Section \ref{sec:sl} and Section \ref{sec:ul}, LSTMs are mostly used for time-dependent sequential data (e.g., action sequences, agent paths, charts for rhythm) and language models, while convolutional neural networks are often used for any type of image-like content. A very popular class of architecture for content generation are GANs~\cite{goodfellow2014GAN}. A GAN consists of two networks, a generator and a discriminator that are trained iteratively to allow the generator to create more realistic content, while the discriminator is getting better at distinguish generated content from real data.}



\subsection{Supervised Learning}\label{sec:sl}

Supervised learning (SL) methods have been used in a variety of ways for content generation. Often as a predictor, SL models predict the gameplay outcomes of games with the generated content, either for evaluating the quality of content, or for meeting specific preferences (such as game style, image style and color) or adapting the generated levels to desired skill-depth. 

\revised{The study of \citet{summerville2016learning} extracted player paths in Mario from gameplay videos and used them to annotate training levels.} Then, separate LSTMs are trained on levels annotated with different players' paths in order to generate personalized levels based on the players' chosen paths \citep{summerville2016learning}. Then, \citet{guzdial2018explainable} trained a random forest on expert-labeled design patterns from Mario levels (i.e., small sections of levels given descriptive class labels) to classify level structures and an autoencoder with level structures and labels as input to generate new instances of those design patterns.

\citet{karavolos2017learning} trained a CNN to predict the outcomes of a simplified 3 versus 3 multiplayer deathmatch shooter game to evaluate and determine if the levels, represented by maps and weapon parameters, are balanced or favor a team. Based on the outcome predictor from~\cite{karavolos2017learning}, \citet{karavolos2018pairing} further designed a DL surrogate model for pairing levels and character classes for desired game outcomes.

\citet{tsujino2018dance} represented rhythm-based video game levels by charts and implemented Dance Dance Gradation (DDG), a system with LSTMs trained on levels of different degrees of difficulty to generate new levels. DDG can tune the difficulty degree of generated charts by changing the fractions of easy or hard charts used to compose the training dataset~\citep{tsujino2018dance}. \citet{liang2019procedural} used C-BLSTM~\citep{schuster1997bidirectional} to generate levels of rhythm games, represented by actions and corresponding timing, of different difficulties, trained on the beatmaps collected from \emph{OSU!}, a famous rhythm game. 

Beside considering skill-depth required in game levels, the emotion sent by content has also been studied. \citet{guzdial2017visual} studied the emotion shown by the game visuals, such as abstract texture, color of game maps and scene, including the visual effects, and trained a CNN to generate textures for some given target emotion.

\citet{soares2019deep} trained a VAE~\cite{Kingma2013vae} to classify NPC behaviors to Leaders, Followers, and Random, in a simple artificial life environment. \citet{sirota2019towards} trained two RNNs, a speaker and a listener, by playing a referential game with concepts and human-generated annotations to design communication systems for NPCs in games. 

\subsection{Standard Unsupervised Learning}\label{sec:ul}
Most unsupervised learning (USL) techniques in PCG focus on learning a representation of all the content and then sample new content from this representation. For example, using autoencoders to learn to replicate game levels. Another direction usually taken is transforming the data into a sequence and use unsupervised learning to learn the relation between these elements similar to Markov Chains relations. For example, learning from a text corpus how to predict the next word based on the previous ones.

\citet{summerville2016super} trained LSTMs on Mario levels annotated with agent paths by representing the 2D levels as one dimensional strings of tiles. \citet{jain2016autoencoders} trained autoencoders on sliding-window segments of Super Mario Bros levels, which were represented by 2D arrays, to generate and repair levels. \citet{jain2016autoencoders} considered a tile as being empty or occupied, but has inspired many follow-up investigations.
Blending has lead to new and creative game levels. \citet{sarkar18blending} trained separate LSTMs on two different game domains (Mario and Kid Icarus), and generated new blended level sections with alternating generators. \citet{sarkar19controllable} further explored generating blended levels by training variational autoencoders and GANs on Mario and Kid Icarus, and generating new blended level sections that interpolate between the domains using the latent vectors. \revised{\citet{snodgrass2020multi} also used VAEs to model and generate platformer level structures which was finished by using a search-based approach to blend details from several other games. \citet{sarkar2020exploring} explored two variants of VAEs (linear are GRU) for blending platforming game levels and associated paths in those levels. \citet{sarkar2020sequential} trained VAEs to learn a sequential model of level segment generation and a random forest classifier to determine the exact location of a newly generated segment to the previous segment (an ancestor). The resulted levels are not only more coherent~\citep{sarkar2020sequential}, but also more creative~\citep{sarkartowards} because of the changing altitude of platformer and various possible heading directions. \citet{yang2020game} trained Gaussian Mixture VAE to learn relation between game level segments from various games (Super Mario Bros, Kid Icarus, and Megaman) and later be able to generate level segments that follow a certain distribution/style. \citet{davoodi2020approach} trained an autoencoder to repair manually designed levels for different games by re-iterating it over the decoder while using a trained discriminator from a GAN model to determine the stopping criteria. Besides levels, autoencoder has also been used to generate 3D shapes~\cite{Yumer2015Procedural}.}

Moreover, USL methods for image generation have also been applied to generating sprites and characters in games. The recent work by \citet{mordvintsev2020growing} learned cellular automata (CA) to imitate the development of organism and generate images, represented by 2D grids of cells. A cell is similar to the tile considered in the MarioGan~\citep{volz2018evolving} (explained later in section~\ref{sec:evocomp}). A cell contains a cell state (e.g., a discrete value or a vector of RGB values), while a tile contains a discrete value which refers to an object type or part of it.

Applications of USL methods to content generation for card games and text adventure games have also been investigated. An example is \cite{summerville2016mystical}.  \citet{summerville2016mystical} trained encoding and decoding LSTMs on Magic: The Gathering cards, represented as sequences of tokens corresponding to the important information on the cards (e.g., mana cost, effect, power, etc.). The LSTMs were trained on corrupted versions of the cards, and encoded cards were used as input to the decoder at generation time.
\revised{Another example is the endless text adventure game \emph{AI Dungeon 2}\footnote{\url{https://github.com/AIDungeon/AIDungeon}} (earlier version as \emph{AI Dungeon}). AI Dungeon 2 is built on OpenAI's GPT-2 model~\citep{radford2019language}, a 1.5B parameter Transformer, and fine-tuned on some text adventures obtained from \url{chooseyourstory.com}, according to its developer Nick Walton~\citep{walton2019dungeon2}.} In a game, a player can interact with the game by inputting text commands, then the AI dungeon master will generate content of the game environment (updates in the game story) according to the commands and provide text feedback.
\revised{By doing so, each player can build his/her own unique game story.
\citet{ammanabrolu2020bringing} focused on procedurally generating interactive fiction worlds and proposed AskBERT to construct knowledge graph. AskBERT extracts objective information in the game worlds, such as characters and objects, via question-answering model.
\citet{ferreira2020computer} proposed \emph{Bardo Composer}, a system that automatically composes music for tabletop role-playing games. In \emph{Bardo Composer}, a BERT model cooperates with a stochastic bi-objective beam search model to identify music emotion, and then generate music pieces that reflects the identified emotion.}

\subsection{Reinforcement Learning}\label{sec:rl}

Using reinforcement learning (RL) for PCG is a very recent proposition which is just beginning to be explored. Here, the generation task is transformed into a Markov Decision Process (MDP), where a model is trained to iteratively select the action that would maximize expected future content quality. This transformation is not an easy task and there is no standard way of handling it. 

One of the early projects that uses RL is by \citet{chen2018q}. They used a small network of one hidden layer to generate a hearthstone deck of cards that can beat a specific other deck given a certain player. The agent can modify the current deck by substituting any of its cards with a different one. The goal is to maximize the win rate of the playing agent using the current deck against a predefined deck. 

\citet{earle2019using} used RL to play the game of SimCity (Maxis, 1989). They used a fractal network (convolutional network with structured skip connections) as their network architecture and optimized it towards increasing the city population. At each step, the agent can change any space on the map to any other type. This project is a borderline example of PCG. The aim of the project was to play the game of SimCity where the trained agent will learn to be a city planner/generator.

\begin{figure}
    \centering
    \begin{subfigure}[t]{0.32\linewidth}
        \centering
        \includegraphics[width=.65\linewidth]{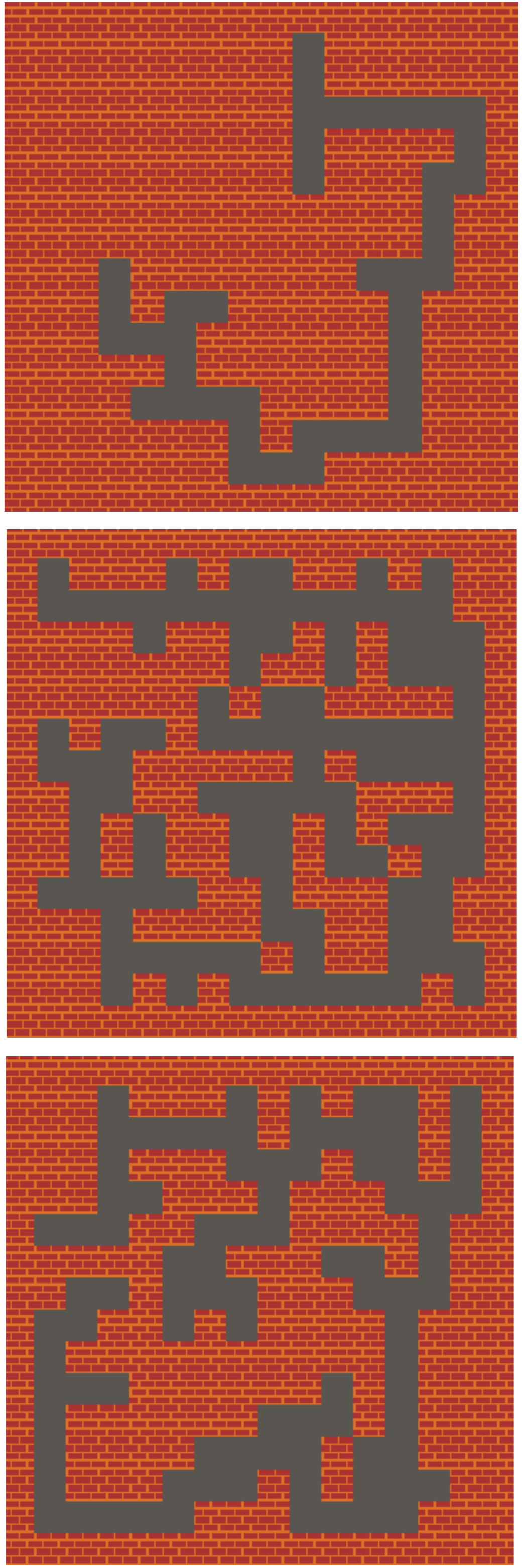}
        \caption{Binary}
        \label{fig:binary_examples}
    \end{subfigure}
    \begin{subfigure}[t]{0.32\linewidth}
        \centering
        \includegraphics[width=.95\linewidth]{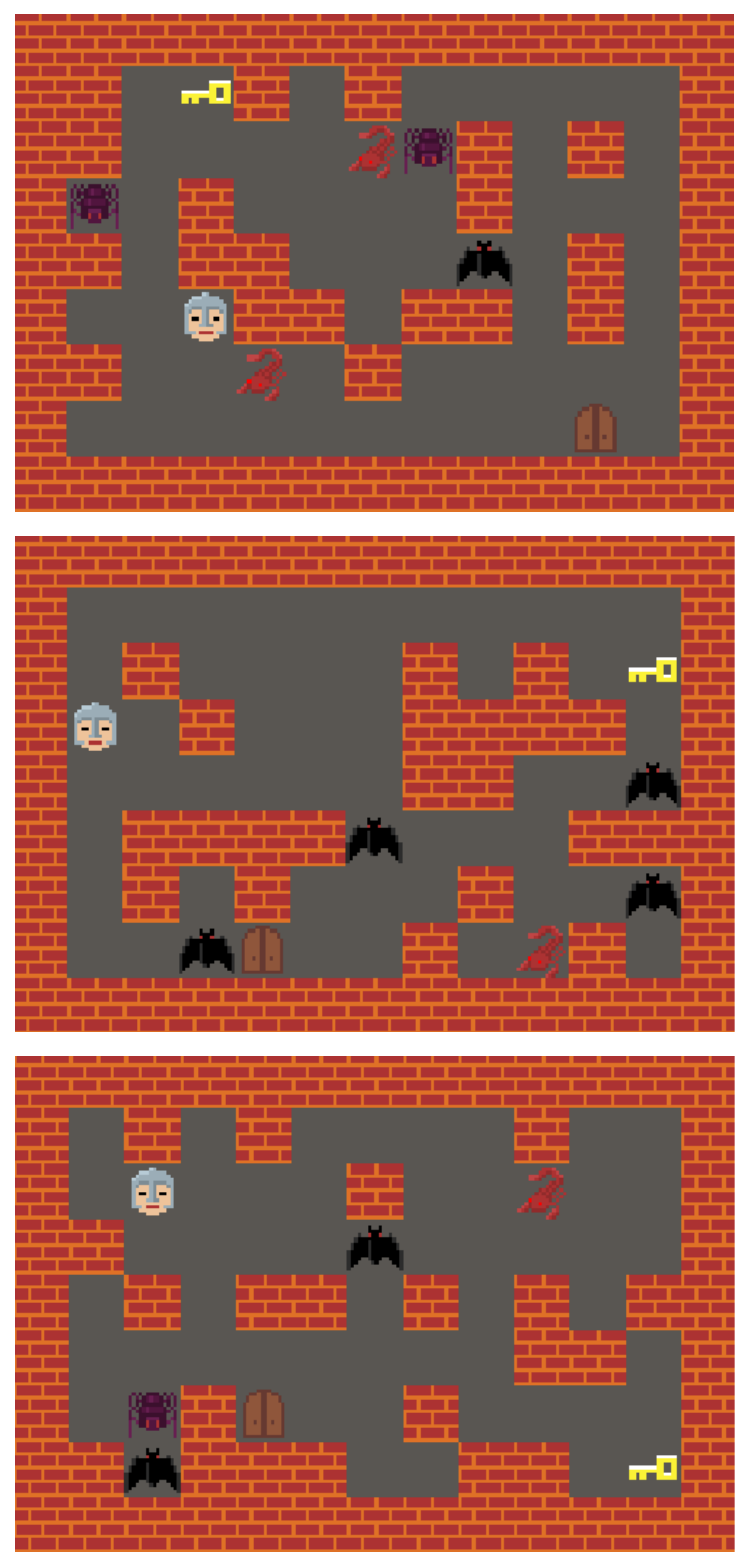}
        \caption{Zelda}
        \label{fig:zelda_examples}
    \end{subfigure}
    \begin{subfigure}[t]{0.32\linewidth}
        \centering
        \includegraphics[width=.65\linewidth]{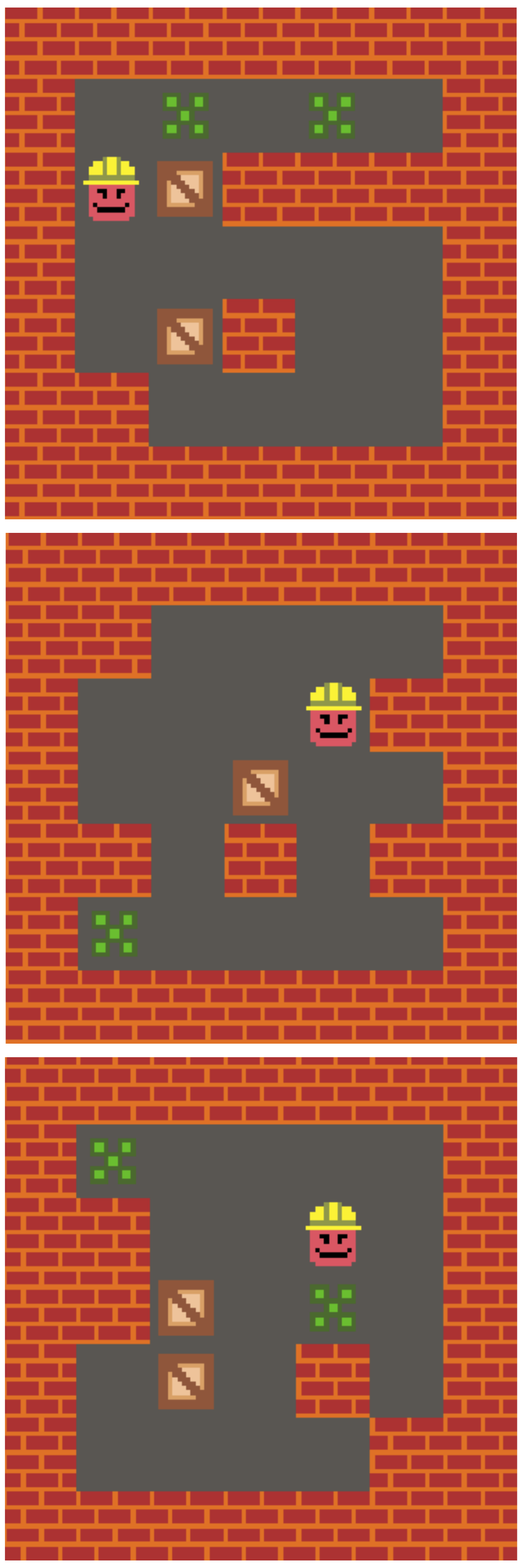}
        \caption{Sokoban}
        \label{fig:sokoban_examples}
    \end{subfigure}
    \caption{Generated examples from three different problems using PCGRL envrionment introduced by 
    \citet{khalifa2020pcgrl}.}
    \label{fig:pcgrl_examples}
\end{figure}

As we can see, most of the RL PCG requires an adaptation for the input to be able to be used during generation. \citet{khalifa2020pcgrl} introduced a framework\footnote{\url{https://github.com/amidos2006/gym-pcgrl}} for 2D level generation using RL. The generation process is framed as an iterative process where at every step the generator modifies the level toward certain goals (based on the current generation problem). They proposed 3 main transformation: Narrow, Turtle, and Wide. These transformation focus on the different ways that the generator controls where it is modifying. \revised{Fig.~\ref{fig:pcgrl_examples} shows examples of the generated levels over three different problems using trained agents in the PCGRL framework.}

\subsection{Adversarial Learning}\label{sec:al}

Adversarial learning (AL) models are perfect for generating content represented by pixel-based images or 2D array of tiles, such as levels as a map, landscapes and sprites. The most popular model among the reviewed works would be GAN~\citep{goodfellow2014GAN} and its variants.

2D levels of most arcade games can be simplified as 2D arrays of tiles, where each tile contains a type of object or part of an object. Examples include the levels designed using Video Game Description Language (VGDL)~\cite{schaul2013pyvgdl} in the General Video Game AI platform~\cite{gvgaibook2019,perez2019general}, and the tile-based levels in the Mario AI framework~\cite{shaker20112010}. As shown in the top-left of Fig. \ref{fig:mariogan}, each tile contains a type of object or part of it, such as ground, pipe, empty and enemy, represented either by a symbol or an integer. \citet{ping2020conditional} implemented an interactive map designing system using different generative models to generate 2D maps, which can be further extended to 3D scenes. \citet{torrado2019bootstrapping} designed a new GAN architecture, Conditional Embedding Self-Attention GAN (CESAGAN), to tackle the low quality and diversity issue of generated 2D levels by traditional GANs, and increased the amount of training data to CESAGAN with a bootstrapping technique. They applied their technique to \emph{Zelda}, a dungeon crawler game from GVGAI~\cite{perez2019general}.

To facilitate the input form for generative models, such as GANs, 3D landscapes are often converted to 2D height map. 
\citet{wulff2017deep} trained a deep convolutional GAN (DCGAN) on digital elevation maps sampled from the Alps dataset to generate 2D height maps as input to Unity for creating 3D landscapes for video games.
\citet{giacomello2018doom} converted each 3D DOOM level to several 2D images, among which a \emph{HeightMap} was used to indicate the 3D information and other were top-down images of the corresponding level. In \cite{giacomello2018doom}, two GANs were trained on human-designed levels, one of which took plain 2D images as input and the other used both the images and some of the extracted features.
\citet{park2019generating} trained a multistep DCGAN, adapted from \cite{volz2018evolving}, to generate levels of an educational game, ENGAGE. The levels were represented by a 2D array of tiles, from a top-down view, during training and creation, and then converted to 3D levels to be used in the game~\citep{park2019generating}. \revised{\citet{volz2020capturing} explored the use of GANs in the context of match-3 levels, attempting to model the local and global structures of those levels.
\citet{awiszus2020toad} proposed token-based oneshot arbitrary dimension generative adversarial network (TOAD-GAN), adapted from SinGan~\citep{shaham2019singan}, trained on a single sample level, to generate tile-based levels. In the work using GANs for level generation that have been reviewed so far, game levels are tackled as image only during training while the constraints for validating levels are not considered at all. Recently, \citet{di2020efficient} presented constrained adversarial networks (CANs) which encourages the generator to learn to generate valid levels by penalizing it due to invalid structures generated during training. But still, these methods generate individual segments of platformer levels separately and then combine them together randomly or according to some increasing level difficulty~\citep{volz2018evolving}. Different from above work, \citet{fontaine2020illuminating} proposed latent space illumination (LSI), which uses quality diversity algorithms, such as Covariance Matrix Adaptation MAP-Elites (CMA-ME)~\citep{Fontaine2020CovarianceMA}, to search the latent space of trained generators, aiming at increasing the diversity of generated levels. A recent work by \citeauthor{Kumaran2020Generating} focused on generating levels in multiple distinct games. Instead of training several GANs for these games separately, a novel GAN architecture, composed of a branched generator and multiple parallel discriminators, was proposed~\citep{Kumaran2020Generating}.
}

\revised{Besides generating 2D and 3D levels represented as pixel-based or tile-based images, texture~\citep{fadaeddini2018case} and sprite generation~\citep{hong2019game} have also been investigated.} \citet{hong2019game} generated 2D image sprites using a multi-discriminator GAN, in which two encoders were used for bone graph, shape and color, without sharing parameters. Additionally, two discriminators, one for shape and the other for color, were used in \cite{hong2019game} to generate sprites' skeletons and color, respectively.
Another potential application is GAN-based character generation~\cite{Jin2017Towards} for video games, such as The Sims (Maxis, 2000). \revised{\citet{wang2020sketch} proposed Sketch2Map to generate 3D terrains from sketches. Sketch2Map used a conditional GAN (cGAN) to convert a sketch into an elevation bitmap, which is interpreted to generate the practical terrain asset by a deterministic algorithm~\cite{wang2020sketch}.}

More recently, \citet{bontrager2020fully} proposed a new training method similar to GANs, where the network consists of two parts: generator and agent. The generator is trying to generate new playable levels adapted to the agent's strength, while the agent plays the game and reports how playable it is and how hard it is to play. Similar to GANs, the agent will try to improve itself by playing the new generated levels, while the generator will improve itself based on the agent performance on its generated levels. In this work, RL is used to play the generated content and not to generate the content; an RL agent interacted with the generative model to create levels adapted to the agent's playing strength.



\subsection{Evolutionary Computation}\label{sec:evocomp}


\begin{figure}
    \centering
    \includegraphics[width=.5\textwidth]{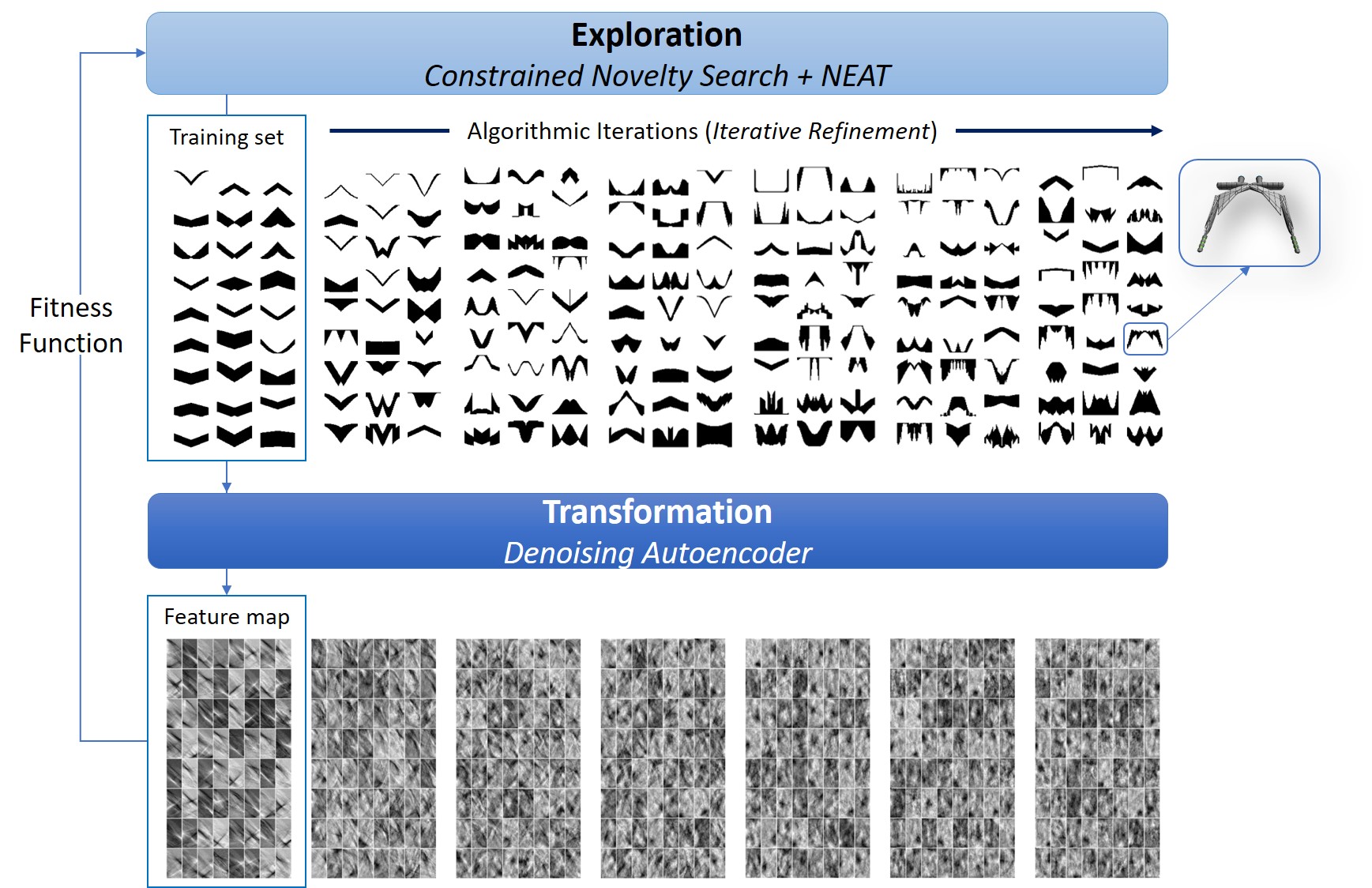}\\
    \caption{\label{fig:delenox}The key phases of DeLeNoX for the autonomous generation of content~\citep{liapis2013transforming}. DeLeNox adopts the principles of exploration (realized via constrained novelty search), transformation (realized via deep denoising autoencoders) and iterative refinement (realized through the increasing complexity of NEAT architectures). Image reproduced with authors' permission.}
\end{figure}

\revised{There is a long tradition of using evolutionary computation (EC) approaches for training (deep) neural networks. While these are sometimes not regarded as DL, the standard definition of DL does in fact not reference gradient descent. Most evolved networks are deep, and architectures created by evolutionary algorithms such as NEAT~\citep{Stanley2002Evolving} often have multiple layers and recurrent components~\citep{Schmidhuber2015Deep}.}

For example, \citet{hoover2015composing} represented game levels as functional scaffolding for musical composition voices~\citep{hoover2014functional}. Taking Mario as an example, each level is presented by a set of voices with the size of possible tile types in a level. Each voice is a one dimensional array of same length of the level, in which each element indicates the vertical position of the tile if it presents on the corresponding column, otherwise 0. Neural networks were trained and evolved through neuroevolution of augmenting topologies (NEAT)~\citep{Stanley2002Evolving} to suggest placements of tiles in Mario levels~\citep{hoover2015composing}.

\citet{hoover2015audioinspace} evolved CPPNs through NEAT for generating both audio and visual content in the game AudioInSpace.
\citet{risi2015petalz} evolved and trained CPPNs with NEAT to generate flower images for a flower-breeding video game  \emph{Petalz}\footnote{\url{https://www.facebook.com/Petalz-238904402867390/}}. The CPPNs of different flowers can be mated to generate new flowers.

\revised{Evolutionary Computation techniques have also been combined with unsupervised DL methods for generating new content. A prominent example is the Deep Learning Novelty Explorer (DeLeNoX)~\cite{liapis2013transforming}.} DeLeNoX alternates phases of content exploration and content transformation for the generation of spaceships, depicted as 2D black and white images (Fig.~\ref{fig:delenox}). In the exploration phase, constrained novelty search seeks maximally diverse artifacts and generates a training set. In the transformation phase, a deep autoencoder learns to compress the variation between the found artifacts into a lower-dimensional space. The newly trained encoder is then used as the basis for a new fitness function, transforming the search criteria for the next exploration phase~\cite{liapis2013transforming}. The process continues repeating exploration and transformation phases thereby iteratively refining and complexifying the generated outcomes.


\begin{figure}
    \centering
    \includegraphics[width=.45\textwidth]{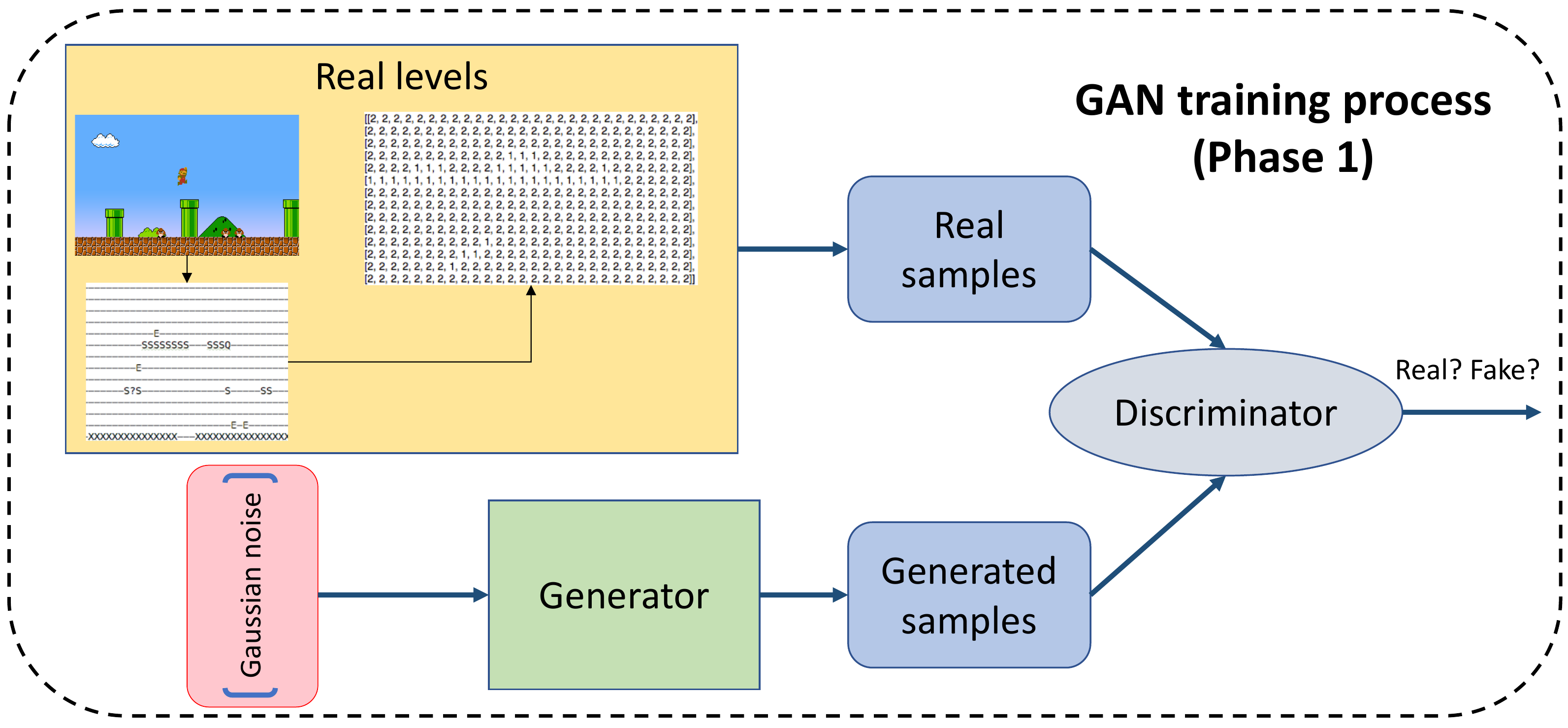}\\
    \includegraphics[width=.45\textwidth]{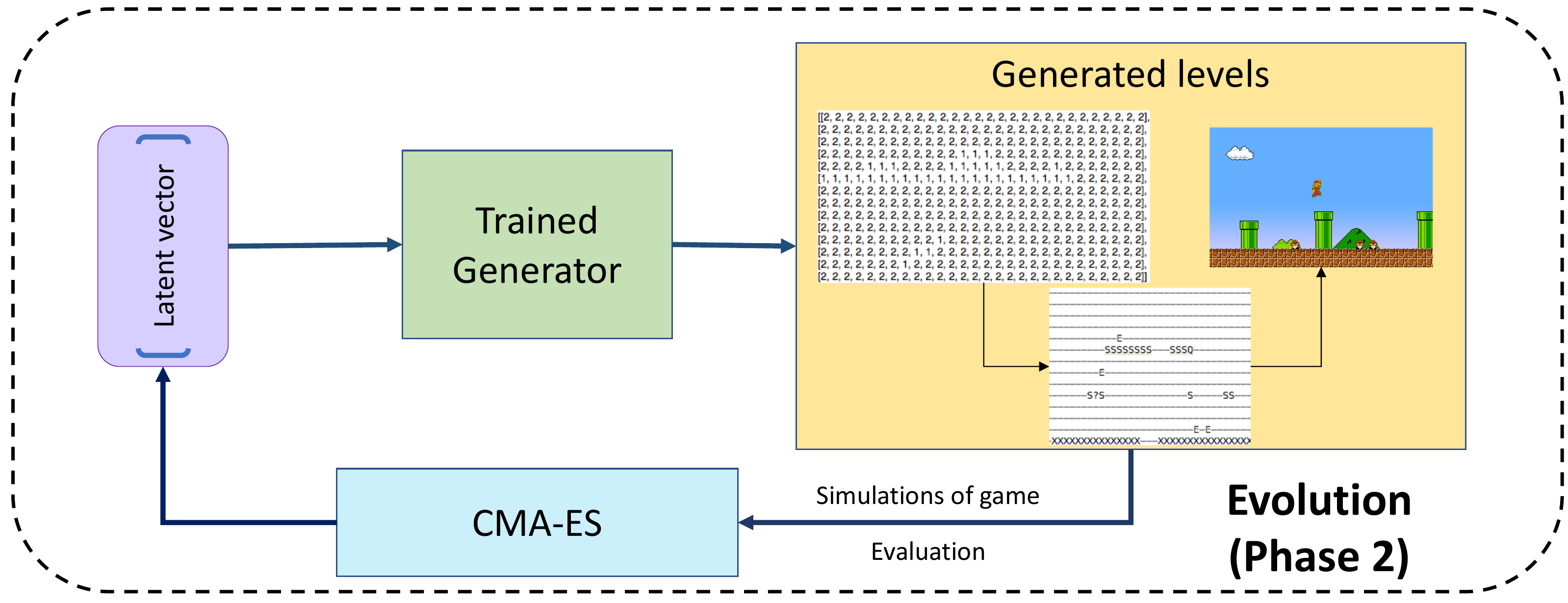}
    \caption{\label{fig:mariogan}Overview process of MarioGan~\citep{volz2018evolving}, reproduced with authors' permission.}
\end{figure}

Arguably one of the most popular examples of EC for DLPCG is the aforementioned Latent Variable approach~\citep{bontrager2018LVE}, which combines unsupervised learning in the form of a GAN/VAE with evolutionary computation to search for content in the learned space of a GAN/VAE. Originating from synthesizing new fingerprint~\citep{Roy2017MasterPrint}, in the context of games this approach has been employed to generate Super Mario Bros and Zelda levels~\citep{volz2018evolving,schrum2020cppn}.

\begin{figure}[htbp]
    \centering
    \includegraphics[width=.49\textwidth]{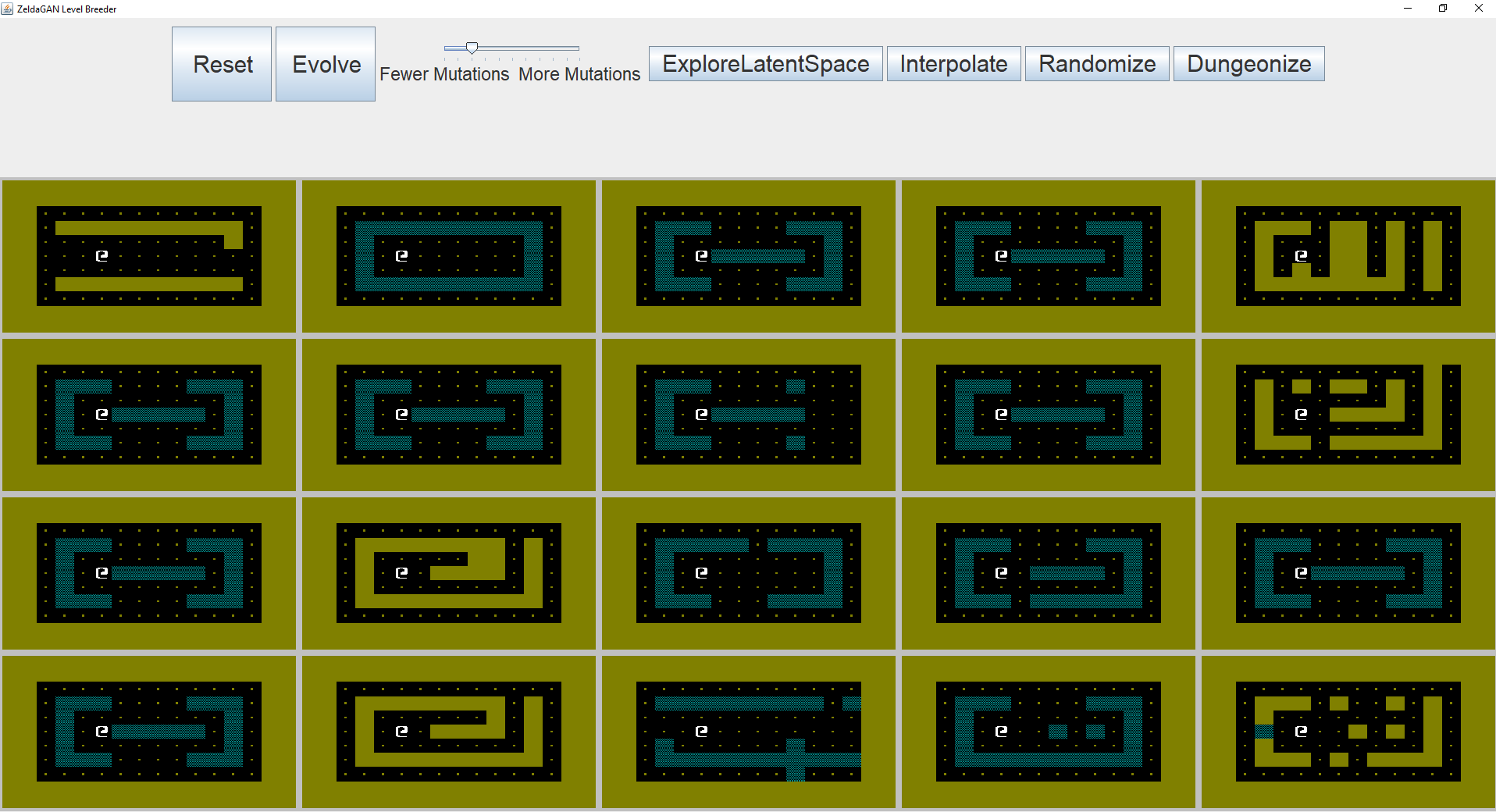}
    \caption{Screenshot of interactive evolution interface in \cite{schrum2020interactive}, reproduced with authors' permission.}
    \label{fig:ganexplorer}
\end{figure}

\begin{figure*}[htbp]
    \centering
    \includegraphics[width=1\textwidth]{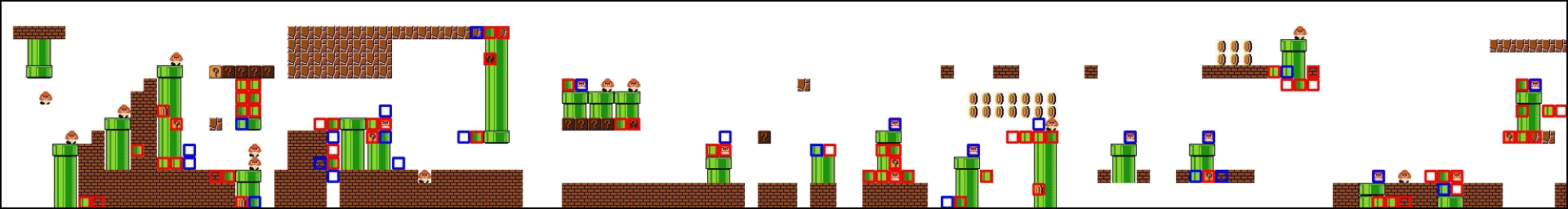}\\
    \includegraphics[width=1\textwidth]{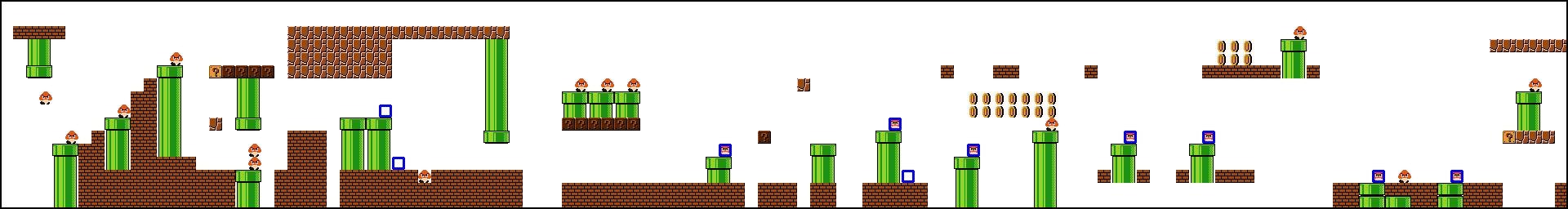}
    \caption{Top: A MLP model trained on human-designed levels labels wrong tiles (in red rectangle) and unsure tiles (in blue rectangle) in a segment. Bottom: Segment fixed by an evolutionary repairer assisted by the trained MLP model~\citep{shu2020cnet}. Images reproduced with authors' permission.}
    \label{fig:pipes}
\end{figure*}

In the work of \citet{volz2018evolving}, a DCGAN~\citep{radford2015unsupervised} is trained on a set of level segments of Super Mario Bros represented by 2D array of tiles, and then latent variable evolution (LVE)~\citep{bontrager2018LVE} is applied to search for levels that are more playable and encourage particular behaviors evaluated by the games simulated by an A* agent. The overview process is illustrated in Fig. \ref{fig:mariogan}. The resulted framework, MarioGAN~\citep{volz2018evolving}, certainly identified a new and creative way of generating game content. However, two issues have been observed: (i) broken pipes occur in some of the level segments generated by GANs, and (ii) the segments were connected directly in an arbitrary order to build complete levels, while how to combine segments to make the resulted levels more structured and organized was not exploited.
\revised{To tackle the former issue, \citet{shu2020cnet} trained a MLP model to learn the surrounding information of tiles and detect wrong tiles in the generated segments (e.g., Fig. \ref{fig:pipes}).} An evolutionary repairer is designed to search for optimal replacement tiles for fixing the broken pipe \citep{shu2020cnet}.
To tackle the latter issue, a graph grammar was used to combine rooms of Zelda generated by a GAN into dungeons~\citep{gutierrez2020zeldagan}, and \citet{schrum2020cppn} proposed CPPN2GAN which used Compositional Pattern Producing Networks (CPPNs) to organize level segments generated by GANs into complete levels.

Inspired by \cite{volz2018evolving}, \citet{irfan2019evolving} applied LVE and trained DCGANs on randomly generated levels of 3 single player games from the GVGAI framework~\citep{perez2019general}, Freeway, Zelda and Colourescape. Based on the work of \cite{volz2018evolving}, \citet{mott2019controllable} designed a new fitness function for CMA-ES as a weighted sum of the number of frames that an action is feasible, the fraction of agents that completed a level and the largest fraction to control the difficulty of generated levels. The weights are evaluated and tuned via the human playing-tests performed on the levels generated using the corresponding fitness function \citep{mott2019controllable}.



Evolutionary methods for content generation can also be combined with user feedback, such as through interactive evolutionary computation (IEC), in which human evaluation is used instead of the fitness evaluation by a simulator.
For example, \citet{hastings2009automatic} used CPPNs to represent weapons in a multiplayer video game Galactic Arms Race\footnote{\url{http://gar.eecs.ucf.edu/}}. The CPPNs are evolved during the game playing with the preferences abstracted from the past playing of players. IEC combined with LVE can allow users to breed their own game levels, such as Zelda and Mario~\citep{schrum2020cppn}. \revised{Based on \cite{volz2018evolving,gutierrez2020zeldagan}, a mixed-initiative tile-based level design tool was implemented by \citet{schrum2020interactive}, which allows human to interact with the evolution and exploration within latent level-design space (interface illustrated in Fig. \ref{fig:ganexplorer}), and to play the generated levels in real-time.}

EC methods can also collaborate with human to generate and evaluate or repair game content. \citet{liapis2013sentient} presented \emph{Sentient World} tool which allows interactions with human designers and generates game maps using Neuroevolution via novelty search. \emph{Sentient World} can generate high resolution maps based on the rough terrain sketches drawn by designers, as well as the iterative refining via selection and editing options opened to designers.

\citet{karavolos2018using} generated levels of a first-person shooter (FPS) game with targeting gameplay outcomes, in which a genetic algorithm is used to generate levels of specific fitness values based on the predicted outcomes by a CNN trained on simulated matches.

\section{Using Deep Learning to Evaluate Content and Content Generators}\label{sec:indirect}

Evaluating content generators is not a trivial task. Much of the ML and DL-based PCG work has focused their evaluations on the generated content, and used those evaluations as proxies for evaluating the generator itself. However, the computational creativity community has identified that in order to get a full picture of the generator (or creative program) the process by which the output content is created should be evaluated as well. \citet{jordanous2012standardised,pease2011impact,colton2008creativity} each propose frameworks and methodologies for evaluating the creativity of the process of a generator. \citet{smith2010analyzing} (later expanded on by \citet{summerville2018expanding}) proposed methods for holistically evaluating a content generation approach, by evaluating large swaths of generated content to get a broader understanding of the generative space of a content generator and its biases within that generative space. \citet{summerville2018expanding} focused on ML-based generators, and proposed approaches for highlighting the shortcomings and strengths of a generator through methodically highlighting generated artifacts (e.g., artifact most similar to an artifact in the training set). 

%
%
%
In this section we survey uses of deep learning for content generation in an \emph{indirect} fashion. In particular, we list studies (cf. Fig. \ref{fig:evaluation}) that consider deep learning for testing or evaluating game content through the analysis of generated content (Section \ref{sec:analysis}), construction of human-like playing bots (Section \ref{sec:playing}), or the construction of reliable models of player experience (Section \ref{sec:experience}). We additionally highlight which of these approaches focus on evaluating the generator itself instead of only the content.

\subsection{Analyzing Content}\label{sec:analysis}
Statistical measures on the generated content and similarity measures based on the content used in training set~(e.g., \cite{lucas2019tile}) can give insight into the generative space of a content generator and its biases within that space. Statistical measures can be used to compare the distribution of generated content to the distribution of the training set~\cite{summerville2018expanding}. Similarity measures can also be specifically designed for this task. For example, \citet{lucas2019tile} compared occurrences of small structures in the generated set to their presence in the training set to measure similarity. 

Many similarity and statistical measures suffer from the same drawback of only measuring what is quantifiable. Recent approaches in deep learning can help avoid this drawback by learning latent semantic features of the content. Recent work has developed approaches to style transfer~\citep{liu2017unsupervised,karras2019analyzing} by traversing the learned latent space of the model, and others have analyzed the learned latent space of their models to find semantic meaning in the features~\citep{abdal2019image2stylegan}. These advances have led to the use of latent space-based distance and similarity measures~\cite{Wong2017retrieval}.  
Leveraging the latent space learned by a model to create similarity measures between pieces of content might allow us to develop more semantically meaningful similarity measures in addition to the statistical measures currently in use. As an indicative example of such a research direction, \citet{isaksen2017semantic} categorized tile-based 2D game levels with semantic hashing based on autoencoders. The proposed approach~\citep{isaksen2017semantic} can be used to categorize the generated level segments or rooms and group the ones sharing similar styles to build a complete game level or dungeon. 

\subsection{Playing Content}\label{sec:playing}




In this section we review methods based on ANNs and DL for reliable playtesting which can be used, in turn, to evaluate game content generators in an indirect fashion. 
Simulated playtesting~\citep{volz2018evolving,holmgaard2014evolving,holmgard2018automated} of generated content can give quick insights into the features of the content and the generative space of the content generator~\citep{smith2010analyzing,summerville2018expanding}. 
\citet{guzdial2016deep} propose the use of deep reinforcement learning agents for simulated playtesting as a way of creating more human-like playtraces. \citet{guzdial2016deep} specifically focus on deep RL agents for Mario, where human-like control is simulated by giving the agent imprecise controls via stochastic effects on actions. Similarly, \citet{min2014deep} designed a goal recognition framework based on stacked denoising autoencoders for open-ended games, which can be used to personalize games for different players according to their actions.

\citet{karavolos2017learning} trained a CNN to predict the outcomes of a simplified 3 versus 3 multiplayer deathmatch shooter game to evaluate and determine if the levels, represented by maps and weapon parameters, are balanced or favoring a team. Based on the predictor for the same deathmatch shooter game, \citet{karavolos2018pairing} further designed a DL surrogate model for pairing levels and character classes for desired game outcomes. \citet{gudmundsson2018human} imitated the behavior of human through SL and performed experimental study on non-deterministic puzzle games \emph{Candy Crush Saga} and \emph{Candy Crush Soda Saga}. A CNN was trained on human player data, and then used to predict the action that human players most likely to select when playing levels that were unseen during training~\citep{gudmundsson2018human}. This approach can be used to measure metrics such as the diversity of actions to evaluate generated new levels. Notice, each of these methods focuses on evaluating the generated artifacts, but can be expanded to more broadly evaluating the generator itself if the results of artifact evaluations are used to stratify the generative space or further explore the biases and capabilities of the generator. 

\subsection{Experiencing Content}\label{sec:experience}

Human user trials and surveys can provide the most useful insight into the less quantifiable (i.e. subjective) features of the content and the generation process, such as the human-perceived quality of the generated content over time. A large volume of studies focus on the use of deep learning for modeling aspects of player experience which can be used, in turn, to evaluate the content that is generated and experienced by the player. Player experience is usually provided as annotated labels (ratings or ranks) or even continuous traces via crowdsourcing. Running user evaluations and crowdsourcing labels of subjective aspects such as experience, however, can be a laborious task which may not be feasible if what is desired is the quick iteration on the generative system. One approach for further leveraging the output of a user evaluation is to treat the user evaluations as features to be learned. \citet{larsson2016content} trained neural networks using NEAT to predict the user rating of user-created StarCraft maps. This approach~\citep{larsson2016content} can be extended to evaluate generated StarCraft maps. 

\revised{Within the platformer genre, a series of studies by \citet{shaker2010towards,shaker2012evolving,shaker20112010} investigate the use of DL models of player experience for the generation of experience-tailored Super Mario Bros levels.} \citet{camilleri2016platformer} view a player's believability as a content generation problem and used various forms of deep networks to infer the mapping between game content, gameplay and believability in a Super Mario Bros variant. The networks of that study predict the degree to which a combination of gameplay behavior and a generated level can be considered believable. \citet{guzdial2016deep} trained a CNN to predict rate of the difficulty, enjoyment and aesthetics of game levels and performed case studies on Infinite Mario Bros, which was further enhanced by the features extracted from search history of an A* agent. Similarly, \citet{summerville2017understanding} used a regression model on a large set of statistical measures to find measures that predict those same human evaluations of Mario levels.
\revised{More recently, \citet{pfaudungeons} proposed deep player behavior modeling (DPBM) with a multi-layer perceptron (MLP) trained on behavioral data and game observation to map game states to action probabilities.} 
All aforementioned approaches can be used, for instance, to evaluate generated levels.


The first application of CNNs for modeling player experience is introduced by \citet{martinez2013learning}. CNNs in that study consider and fuse the content of a 3D maze prey-predator game and the in-game behavior of the player \citep{martinez2014deep} and predict reported ranks of player experience via use deep preference learning. Looking at the challenge of player affect modeling by solely focusing on gameplay, \citet{makantasis2019pixels} used various CNN models to predict the level of arousal of survival shooter games directly from the pixels of gameplay in a general player-agnostic fashion. Thus CNNs map between gameplay behavior and game content as represented by pixels---such as in-game play features and UI elements. In principle, such surrogate models of arousal can be used directly and evaluate video content of any game within the the survival shooter genre. In a similar recent study various types of neural networks have been trained to predict the continuous viewer engagement of PUBG streamed games on Twitch \citep{melhart2020moment}; the engagement models obtained are highly accurate and general across different streamers. \citet{camilleri2017towards} took player experience modeling to the next level and built models that are general across many different games. The models are build on simple 1-hidden layer networks indicating the potential of the methodology with larger DL representations for the general evaluation of the experience of game content across games. Similar to the previous section, each of these methods are predominantly used to evaluate content. However, using these methods to evaluate large samples of content from a generator can enable a meta-analysis of the types of content a particular generator tends towards creating.

\begin{figure}[htbp]
    \centering
    \def\angle{0}
\def\radius{3}
\definecolor{clr1}{RGB}{215,48,39}
\definecolor{clr2}{RGB}{244,109,67}
\definecolor{clr3}{RGB}{253,174,97}
\definecolor{clr4}{RGB}{254,224,144}
\definecolor{clr5}{RGB}{255,255,191}
\definecolor{clr6}{RGB}{224,243,248}
\definecolor{clr7}{RGB}{171,217,233}
\definecolor{clr8}{RGB}{116,173,209}
\definecolor{clr9}{RGB}{69,117,180}

\def\cyclelist{{"clr1","clr3","clr5","clr9"}}
\newcount\cyclecount \cyclecount=-1
\newcount\ind \ind=-1
\begin{tikzpicture}[nodes = {font=\sffamily}, scale=0.7]
    \draw[ultra thick,color=black!80,fill=clr1!50] (-5,5) rectangle (0,3.5);
    \draw[ultra thick,color=black!80,fill=clr9!50] (0,5) rectangle (5,3.5);
    \draw[ultra thick,color=black!80,fill=clr5!80] (-5,3.5) rectangle (5,1.5);
    \draw (-2.5,4.5) node(a1) {Analyzing content};
    \draw (-2.5,4) node(a2) {\citep{liu2017unsupervised,karras2019analyzing,abdal2019image2stylegan,isaksen2017semantic}};
    \draw (2.5,4.5) node(b1) {Playing content};
    \draw (2.5,4) node(b2) {\cite{guzdial2016deep,min2014deep,karavolos2017learning,karavolos2018pairing,gudmundsson2018human}};
    \draw (0,3) node(c1) {Experiencing content}; 
    \draw (0,2.5) node(c2) {\cite{larsson2016content,shaker2010towards,shaker2012evolving,shaker20112010,camilleri2016platformer,guzdial2016deep}};
    \draw (0,2) node(c3) {\cite{summerville2017understanding,pfaudungeons,martinez2013learning,martinez2014deep,makantasis2019pixels,melhart2020moment,camilleri2017towards}};
\end{tikzpicture}
    \caption{Summary of the works that focused on analyzing, playing or experiencing generated content.}
    \label{fig:evaluation}
\end{figure}





\def\nouse{
\begin{table}[htbp]
    \centering
    \caption{Open-source PCG frameworks. \jialin{Jialin: we can remove this table}}
    \label{tab:platforms}
    \begin{tabular}{l|l}
    \noalign{\smallskip}\hline
      Platform & URL \\
      \noalign{\smallskip}\hline\noalign{\smallskip}
      MarioGAN+GameGAN~\citep{volz2018evolving,schrum2020cppn} & \url{https://github.com/schrum2/GameGAN}\\
      Graph+GAN~\citep{gutierrez2020zeldagan} & \url{https://github.com/schrum2/MM-NEAT}\\
      PCGRL~\citep{khalifa2020pcgrl} & \url{https://github.com/amidos2006/gym-pcgrl}\\
      GPN~\citep{bontrager2020fully} & \url{https://github.com/pbontrager/GenerativePlayingNetworks}\\
    \noalign{\smallskip}\hline
    \end{tabular}
\end{table}
}
 



\section{Discussion and Outlook}\label{sec:outlook}


The combination of deep learning and PCG in games is beneficial for both game research---as deep learning enhances our capacity to generate content---and deep learning research since games pose challenging problems for deep learning to solve. Deep learning opens new opportunities for the autonomous generation of content of any type and has a plethora of use cases within games. As we saw throughout this article, deep learning may serve as a content generator, as a content evaluator, as a gameplay outcome predictor, as a driver of search, and as a pattern recognizer for repair and style transfer. This section surveys the areas with a particular importance for the current and future use of DLPCG in games with an emphasis on mixed-initiative generation, style transfer and breeding, underexplored content types, learning from small datasets, orchestrating different content types within a game, and generalizing generation across games.

\subsection{Mixed-initiative DLPCG}

Autonomous PCG systems, including the cases where the initiative of the human designer is limited to algorithmic parameterizations~\citep{yannakakis2018artificial}, can hardly generate content with target quality or features. Recently, more and more work takes into account the preferences or input of designers or players in different ways while generating content. Mixed-initiative PCG \citep{Yannakakis2014Mixed}, formally defined as ``the process that considers both the human and the computer proactively making content contributions to the game design task'' \citep{yannakakis2018artificial}, offers a more controllable and practical design process that may involve the use of DLPCG algorithms but their use is limited so far.

Level generation in games, as a popular application of mixed-initiative DLPCG, requires some initial specifications (i.e. the initiative) from the designer---e.g. in the form of  sketches~\citep{ha2017sketch}---to assist the design process. \revised{A popular example of the mixed-initiative paradigm is the shallow neural network model presented in \citep{Liapis2013sentientsketchbook} which generates game strategy maps based on the terrain sketches drawn by designers. The map generation feature of Sentient Sketchbook features neuroevolutionary search which is driven by design objectives and the novelty of the map.} Moving from level to image generation, \citet{serpa2019towards} adapted the GAN-based Pix2Pix architecture to generate both gray and color pixel art sprites from sketches using a single network.

Taking platform games as the domain under investigation, \citet{guzdial2018co} developed a mixed-initiative Super Mario Bros level design tool that leveraged several existing PCGML techniques, including Markov chains~\citep{snodgrass2014experiments}, LSTM~\citep{summerville2016super} and Bayes Net~\citep{guzdial2016game}, to assist the user in creating levels. \citet{guzdial2018co} gathered data on how the users interacted with the models in the tool, and trained a CNN on that collected data. This CNN was then used to better predict and generate level sections along with the user. 
Later, \citet{guzdial2019friend} used the trained CNN with active learning based on the user current interaction to generate levels for Super Mario in a mixed-initiative fashion~\citep{Yannakakis2014Mixed,Liapis2016Boosting}.
Recently, \citet{schrum2020interactive} allowed the designers to change manually the latent vectors of the trained generative model or define the mutation strength of their evolutionary generator for tile-based 2D levels. 
\revised{\citet{delarosa2020mixed} presented \emph{RL Brush}, a human-driven, AI-augmented design tool also for tile-based 2D levels, in which RL-based models have been used to enhance human design with suggestions generated by PCG methods.}




\subsection{Style Transfer, Breeding and Blending}
Most style transfer methods and generative models for image, music and sound~\citep{briot2019deep}, can be applied to generate game content. So far, only a few work focused on the style transfer for game content~(e.g., \cite{yoo2016changing,liapis2013sentient,serpa2019towards}). 
\cite{liapis2013sentient} generated game maps based on the terrain sketches and \cite{serpa2019towards} generated art sprites from sketches drawn by human. However, a number of diverse input sketches to these two work can also be generated using deep learning approaches based on a single human sketch~\cite{ha2017sketch}.
Moreover, algorithms and techniques designed for image generation can often be adapted to the automatic generation of faces and sprites in games. For instance, \cite{yoo2016changing} applied a neural styling algorithm~\cite{gatys2015neural} to change artistic style of graphics in a strategy game \emph{Hedgewars}\footnote{\url{http://www.hedgewars.org/}}.
\revised{Another example is \emph{ArtBreeder}\footnote{\url{https://artbreeder.com/}}, which contains several generative models for creating new images by image breeding, among which, the models for portraits and anime-style faces, can be used to generate comic or video game characters and the one for landscapes can be used to generate background images for games. 
Blending levels from different games has recently gained more attention from the research community, with much recent work focusing on blending platformer levels. \citet{sarkar18blending} and \cite{sarkar19controllable} trained separate models on two different games, and then blended new levels using these trained models via interpolation or alternation. \citet{snodgrass2020multi} used VAEs to generate level structures, and a search-based approach to blend details from various platformers, while \citet{sarkar2020exploring} directly trained VAEs on levels from several platforming games and interpolated the latent vectors between domains for blending.}

\subsection{Underexplored Content Types}

Most of the reviewed works focus on the design of content that can be represented by 2D images of tiles or pixels, such as 2D levels, landscapes and sprites (cf. Section \ref{sec:methods}). 
Only a few of them considered text and narrative generation, music and rhythm generation, weapons generation for FPS, etc. 

\revised{In the research we have surveyed, platformer and dungeon-like games (e.g., arcade games, FPS games and adventure games) are clearly over-represented. In particular, Super Mario Bros and Zelda are usually used for testing the GAN-based level generation approaches.}

However, the types of games are not limited to arcade games and the generation of some commonly seen types of game content are rarely investigated. For instance, the generation of characters (skills, actions, and images) for fighting games and multi-player online battle games; the generation of cards and rules for strategy card games (e.g., Hearthstone); event generation (stories and effects) (e.g., for The Sims); goal generation in all kinds of games.
Several approaches from other fields can be adapted to DLPCG, such as transfer learning for image generation in games, story generation for text-based adventure games and conversational NPCs.


\subsection{Content Generation in Real-time - Personalized Game content}

Another less explored area is content generation in real-time, such as generating level segments during gameplay, according to the actual player's playing skill-depth, style and preferences. Taking Super Mario Bros as an example, several MarioGAN models can be trained offline using a variety of fitness functions with different aims (e.g., encourage more jumps by putting more pipes, put more coins \revised{for players to collect}, adjust the difficulty by controlling the number of enemies), and then be selected to generate new level segments during the game after determining the player's preferences and performance according to the gameplay data during first segments.

\subsection{Learning from Small Data}
One of the main limitations for most forms of PCG based on deep learning, or PCGML in general, is the access to training data. Some games have a large amount of existing content, either made by developers or by users. However, for a game in development there may not be content to learn from, because the content may not be made yet. In fact, not having to produce all of that content may be a prime reason for wanting to train a content generator in the first place. What would be desirable here would be a way of training a generator based on only a few pieces of hand-designed content, such as items, levels, or characters.

One approach to doing this is \emph{bootstrapping}, where a generator is first trained on just a few examples, and whenever it produces new content that satisfies the functionality constraints, this content gets added to the training set for continued training of the generator~\cite{torrado2019bootstrapping}. This approach requires a reliable test of the functionality constraints, for example the playability of a level can be tested with game-playing agents.

Note that the amount of data required to train a reliable model varies greatly depending on the complexity of the model, the complexity of the data, and the training procedures of the model. For example, the training data limitation does not apply to PCG methods based on reinforcement learning. Further, MarioGAN~\cite{volz2018evolving} was trained on a single Mario level broken into many sections. \citet{snodgrass2017studying} explored the effects of the amount and diversity of training data on a simple Markov chain model and an LSTM, and found that the benefits of additional data dropped off after several levels. Further studies exploring the data requirements of DLPCG models can help illuminate the usability and scalability of these approaches.

\subsection{Generalization across Games}

Another, and arguably better, approach to learning generators for games for which you do not (yet) have much content would be trained on content from other games. After all, games from a particular genre have much in common, and it should arguably be possible to train on FPS levels from \emph{Quake}, \emph{Halo} and \emph{Call of Duty} to learn to generate new levels for \emph{Half-Life}. It should be even easier to \revised{train character models on existing human-designed characters from several open-world games,} as they share the same functionality constraints. The trained generator would likely be a conditional model, that takes some encoding of the characteristics of a game as input. In all of these cases, the deep learning model would have to learn to represent the underlying similarities between content for the games it was trained on, as well as the differences.


\subsection{Orchestration for Game Generation}

A key future research direction for any PCG framework is the generation of more than one domain of computational creativity within games. The six key computational game creativity domains as defined by \citet{liapis2014computational} include visuals, audio, narrative, levels, rules and gameplay. A process that considers the output of two or more of these domain generators up to the generation of a complete game is referred to as \emph{orchestration} \cite{liapis2018orchestrating}. In other words, orchestration can be defined as the ``harmonization of the game generation process'' \cite{liapis2018orchestrating}. 

While orchestration is a core aim for the autonomous generation of complete games \citet{liapis2018orchestrating} reported only a few game generation systems that considered more than one generation domain. These include Angelina \cite{cook2016angelina,cook2013mechanic}, Game-O-matic \cite{treanor2012game}, Sonancia \cite{lopes2015sonancia}, AudioInSpace \cite{hoover2015audioinspace} and the FPS generator by \citet{karavolos2018pairing,karavolos2019multi}. Among these case studies of orchestrated game generation only a few can be considered early embryos of DLPCG-based game orchestration. In particular, the work by Karavolos et al. \cite{karavolos2018pairing,karavolos2019multi}, Sonancia \cite{lopes2015sonancia}, and AudioInSpace \cite{hoover2015audioinspace} use various forms of shallow and deep neural networks---both as surrogate models (indirectly) and as generative functions (directly)---to generate content for multiple domains within games. As deep learning is of particular importance for fusing the generation process across content representations of dissimilar resolutions and characteristics \cite{yannakakis2018artificial}, we expect to witness an increase in DL research work towards achieving game orchestration.




\section{Conclusions}



\revised{The work surveyed in this paper is the result of two convergent trends from the last few years. One is the increasing use of deep learning for generative tasks in non-game contexts, such as GANs and VAEs used for generating pictures of faces and RNNs used for generating voices and music. The other is the increasing use of machine learning in PCG, something that was unheard of until five years or so ago. Both of these trends build on the deep learning revolution itself, which has made machine learning effective on completely new classes of problems.}

\revised{As a result, interest in deep learning for PCG has exploded. Examples abound, as our survey shows. It is very likely that we will see rapid progress in this research direction in the near future. This survey paper attempts to contribute to this progress by surveying and systematizing this work and implicitly and explicitly pointing out relevant and fertile research problems. We believe that this is a very timely effort given the exciting pace of this field.}



\revised{Deep learning methods have been applied alone or in collaboration with other PCG methods to generate game content and to analyze, play and experience content. Due to the characteristics of different types of content, different types of deep neural architectures have been used. Among the reviewed work, the widely used neural architectures include convolutional neural networks for supervised learning tasks, varying from generating texture or music for target emotion to predicting game outcomes or difficulty rate; long short-term memory for generating sequential data like charts for rhythm and narrative or for predicting action sequences; deep variational autoencoders, mostly used for generating level maps and sometimes for classifying NPCs' or players' behaviors; and generative adversarial networks for creating image-like content (e.g., level maps, landscapes, faces and sprites).
A part from the direct use of deep learning methods or their alliance with evolutionary computation to generate game content, they have also been used for evaluating content and content generators in an indirect manner.}

\revised{Although a variety of game content (e.g., levels, text, character models, textures, music and sound) have been investigated, the generation of content like event, goals or character features with skill-depth can be exploited more. As a future research, evolving or training game-playing agents and content generators in parallel, such as in the recent work of \citet{dharna2020coevolution}, is of great interest, as well as the generalization across games. Besides those, online generation of game content to adapt players' skill and preferences in real-time will accelerate the realization of personalized games.}

\begin{acknowledgements}
J. Liu was supported by the National Natural Science Foundation of China (Grant No. 61906083), the Guangdong Provincial Key Laboratory (Grant No. 2020B121201001), the Program for Guangdong Introducing Innovative and Entrepreneurial Teams (Grant No. 2017ZT07X386), the Science and Technology Innovation Committee Foundation of Shenzhen (Grant No. JCYJ20190809121403553), the Shenzhen Science and Technology Program (Grant No. KQTD2016112514355531) and the Program for University Key Laboratory of Guangdong Province (Grant No. 2017KSYS008).
S. Risi was supported by a Google Faculty Research award and a Sapere Aude:DFF-Starting Grant. 
A. Khalifa and J. Togelius acknowledge the financial support from National Science Foundation (NSF) award number 1717324 - ``RI: Small: General Intelligence through Algorithm Invention and Selection''. G. N. Yannakakis was supported by European Union's Horizon 2020 AI4Media (951911) and TAMED (101003397) projects. 
This is a pre-print of an article published in Neural Computing and Applications. The final authenticated version is available online at: \url{https://doi.org/10.1007/s00521-020-05383-8}.
\end{acknowledgements}

%
 \section*{Conflict of interest}
S. Snodgrass, S. Risi, G. N. Yannakakis, and J. Togelius declare that they a financial interest in \emph{modl.ai}, which develops AI technologies for games.
\bibliographystyle{spbasic}      
\bibliography{main}   
\balance
\end{document}